\documentclass{article} 

\usepackage{xcolor}
\usepackage{arxiv}

\usepackage{tcolorbox}
\usepackage{booktabs}
\usepackage{multirow}
\usepackage{hyperref}       
\usepackage{url}            
\usepackage{amsfonts}       
\usepackage{nicefrac}       
\usepackage{microtype}      
\usepackage{doi}
\usepackage{amsmath}

\usepackage{multirow}
\usepackage{stix} 
\usepackage{mathtools}
\usepackage{enumitem} 

\usepackage{tcolorbox}
\usepackage{hyperref}
\usepackage{amsmath}

\definecolor{darkgreen}{rgb}{0,0.5,0}

\usepackage{subcaption}          
\usepackage[utf8]{inputenc}      
\usepackage[T1]{fontenc}         
\usepackage{caption}
\usepackage[normalem]{ulem}

\newcommand{\colorsquare}[1]{\textcolor{#1}{\textbullet}}

\definecolor{c_female}{RGB}{250, 252, 108}
\definecolor{c_male}{RGB}{48, 242, 49}
\definecolor{c3}{RGB}{111, 248, 244}

\AtBeginDocument{%
  }

\title{\textit{Aesthetics as Structural Harm:} Algorithmic Lookism Across Text-to-Image Generation and Classification}
\renewcommand{\shorttitle}{Aesthetics as Structural Harm}

\author{ Miriam Doh \\
        Université Libre de Bruxelles \\
        Brussels, Belgium \\
        \texttt{miriam.doh@umons.ac.be} \\
        \And
        Aditya Gulati\\
	    ELLIS Alicante\\
        Alicante, Spain \\
	    \texttt{aditya@ellisalicante.org} \\
        \And
        Corinna Canali \\
        Universität der Künste Berlin and \\Weizenbaum Institute \\
        Berlin, Germany \\
        \texttt{corinna.canali@drlab.org}
        \And
        Nuria Oliver\\
	    ELLIS Alicante\\
        Alicante, Spain \\
	    \texttt{nuria@ellisalicante.org} \\
}

\begin{document}

\maketitle

\begin{abstract}
This paper examines \textit{algorithmic lookism}—the systematic preferential treatment based on physical appearance—in text-to-image (T2I) generative AI and a downstream gender classification task. Through the analysis of 26,400 synthetic faces created with Stable Diffusion 2.1 and 3.5 Medium, we demonstrate how generative AI models systematically associate facial attractiveness with positive attributes and vice-versa, mirroring socially constructed biases rather than evidence-based correlations.
Furthermore, we find significant gender bias in three gender classification algorithms depending on the attributes of the input faces. Our findings reveal three critical harms: (1) the systematic encoding of attractiveness-positive attribute associations in T2I models; (2) gender disparities in classification systems, where women's faces, particularly those generated with negative attributes, suffer substantially higher misclassification rates than men's; and (3) intensifying aesthetic constraints in newer models through age homogenization, gendered exposure patterns, and geographic reductionism. These convergent patterns reveal algorithmic lookism as systematic infrastructure operating across AI vision systems, compounding existing inequalities through both representation and recognition.

\textbf{Disclaimer:} \textit{This work includes visual and textual content that reflects stereotypical associations between physical appearance and socially constructed attributes, including gender, race, and traits associated with social desirability. Any such associations found in this study emerge from the biases embedded in generative AI systems—not from empirical truths or the authors’ views. }
\end{abstract}


\section{Introduction and Related Work}
Text-to-image generative AI systems have become increasingly prevalent in shaping online visual content \cite{ricker2024ai,Karagianni14112024,
ricker2024ai,epstein2023art}, creating new forms of algorithmic harms that extend far beyond traditional notions of algorithmic bias \cite{bender2021dangers,weidinger2021ethical}. While the rapid adoption of these models promises democratized content creation, they simultaneously generate systematic disadvantages for marginalized communities through mechanisms that are often 
invisible to both users and developers \cite{birhane2021multimodaldatasetsmisogynypornography}. 
This widespread adoption has intensified concerns about detecting, measuring, and addressing biases that these systems may encode and amplify \cite{Hall2022}, revealing that generative AI systems do not merely reflect existing societal prejudices but amplify and institutionalize them through \textit{synthetic normativity} \cite{startari2025ethos}, \emph{i.e.}, the algorithmic construction of aesthetic and social norms that determine who becomes visible and valued in AI-generated content and on AI-moderated platforms.

Existing research has extensively documented gender \cite{Wang2019, Wang2020,Schwemmer2020}, racial \cite{Yucer2020,Howard2024,Khan2021}, and age-based \cite{Karkkainen2021,JacquesJunior2019} disparities in computer vision and generative image systems, including T2I models. These models have been shown to overrepresent whiteness and masculinity while marginalizing other demographics \cite{25_luccioni2023stable,26_naik2023social}. Such disparities extend beyond identity categories to encompass object associations, attire representation, and spatial positioning \cite{wu2023stable}, as well as subtler forms of discrimination identified by Kumar \emph{et al.} \cite{Kumar2024}, including the \textit{representative bias} (\emph{i.e.}, the disproportionate privileging of the perspectives or experiences of particular identity groups), and the \textit{affinity bias} (\emph{i.e.}, the systematic preference for particular narratives or viewpoints). While examined initially in the context of language 
models, these dynamics can also emerge in generative image systems, as illustrated by Qadri \emph{et al.} \cite{qadri2023ai}, who document how visual outputs encode skewed demographic portrayals and culturally biased aesthetics, even when prompts are demographically neutral.
\begin{figure}[h]
    \centering
    \includegraphics[width=0.9\linewidth]{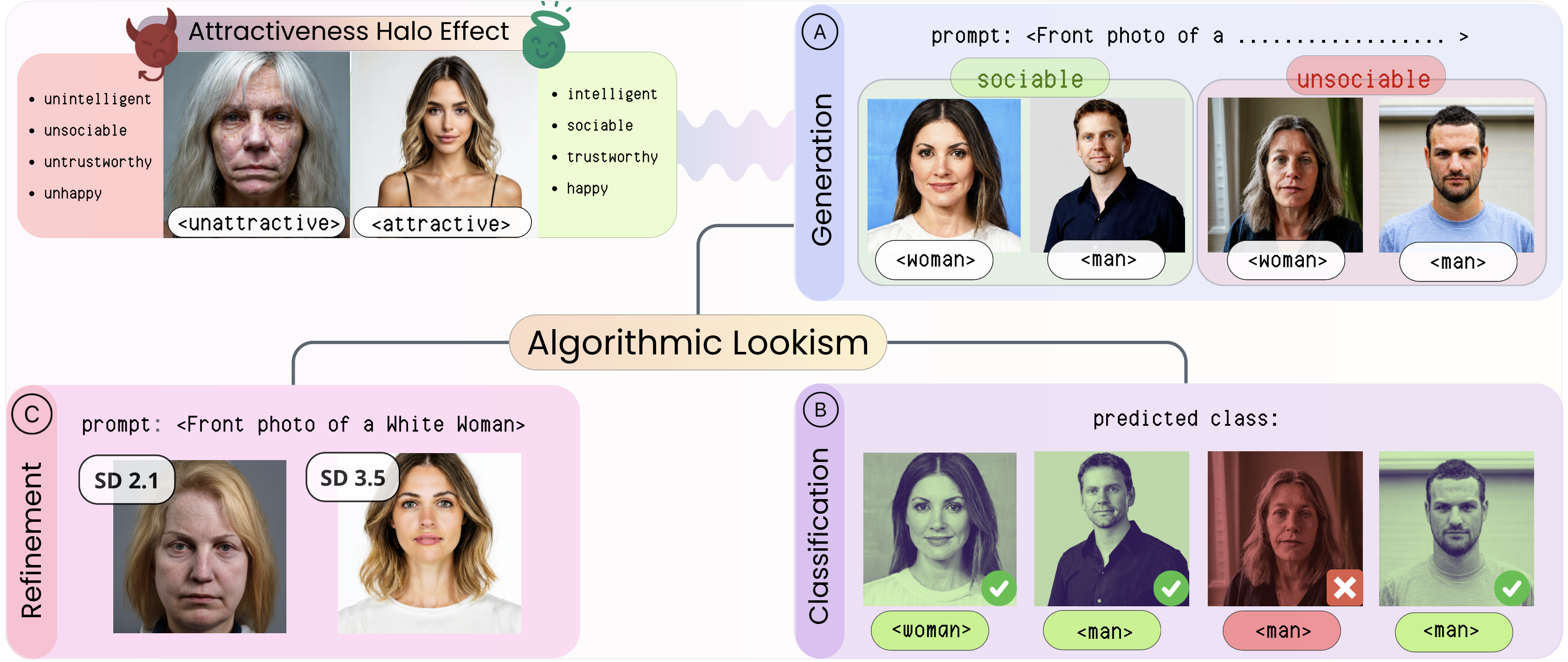}
    \caption{The attractiveness halo effect (top) guides dataset generation to 
test algorithmic lookism across \textbf{(a)} generation, \textbf{(b)} classification, and 
\textbf{(c)} aesthetic model refinement in SD 3.5 vs. 2.1. Images in the example are for images generated of White women and men.}

    \label{fig:exsd35}
\end{figure}

\textit{Algorithmic lookism} \cite{gulati2024lookism}, \emph{i.e.}, the systematic preferential treatment of individuals based on their physical appearance, constitutes a particularly insidious form of structural harm. Grounded in beauty ideals and psychological biases \cite{Dion1972, Talamas2016, Eagly1991, Tversky1974, gulati2024beautifulgoodattractivenesshalo}, algorithmic lookism creates cascading disadvantages that go beyond representation.

In fact, the harms caused by algorithmic lookism operate across multiple AI vision systems. Image generation and analysis models exhibit parallel \textit{``politics of (in)visibility''}: Western-centric aesthetic norms determine whose faces become visible in generation and accurately classified in classification. These appearance-based biases reflect market-driven optimization that reconfigure human identity into profitable data, with synthetic images potentially amplifying these patterns when used to train downstream systems.

In this paper, we examine how algorithmic lookism operates systematically 
through the correlation of facial attractiveness with positive attributes in 
images generated using Stable Diffusion 2.1 (SD 2.1) and 3.5 Medium (SD 3.5).  
 
Extending prior work on SD 2.1 \cite{pmlr-v294-doh25a}, we broaden the analysis through: (1) comparative evaluation of SD 2.1 vs. 3.5 to assess whether enhanced data curation reduces or intensifies aesthetic bias, (2) qualitative analysis documenting aesthetic patterns beyond quantitative metrics, and (3) feminist-informed critique situating algorithmic lookism within neoliberal rationality and colonial logics.

Specifically, we examine four attributes--namely happiness, intelligence, sociability, and trustworthiness--which Gulati \emph{et al.} \cite{gulati2024beautifulgoodattractivenesshalo} employed in their empirical investigation of attractiveness-based perception with human participants, in their positive and negative versions. These attributes draw from an extensive social psychology literature \cite{Dion1972,Kanazawa2004,Talamas2016,Mathes1975,Golle2013,Todorov2008,Miller1970} documenting systematic perception biases whereby physical attractiveness becomes associated with positive traits and behavioral characteristics, despite limited empirical evidence for such correlations.

Adopting these same definitions enables a systematic comparison between our 
findings from AI-generated imagery and Gulati \emph{et al.}'s empirical results with human subjects, supporting methodological consistency across studies.

Furthermore, we evaluate the impact of algorithmic lookism on computer vision-based gender classification models not merely as a downstream application, but as evidence of pervasive aesthetic normativity operating 
consistently across AI systems. Building on work by Doh \emph{et al.} \cite{doh2024my}, we treat differential classification performance as diagnostic evidence that aesthetic standards function as systematic filtering mechanisms determining who becomes \textit{algorithmically readable} in digital identity systems.

This comparative analysis reveals algorithmic lookism as a systematic infrastructure: not an isolated technical failure but part of a coherent normative framework that determines both visibility in generation (Figure \ref{fig:exsd35}.(a)) and legibility in classification (Figure \ref{fig:exsd35}.(b)). More concerning, SD 3.5 exhibits intensified aesthetic constraints despite reported improvements in data curation, with enhanced filtering appearing alongside systematic narrowing of diversity and alignment with market-driven 
aesthetic hierarchies (Figure \ref{fig:exsd35}.(c)).

\section{Methodology}
In this section, we describe the dataset used across all analyses, formulate our research questions, and detail the methodological approaches employed to address each question.
\subsection{Dataset Creation}
\label{sec:methods.dataset}
Two different datasets, each comprising 13,200 face images, were generated by prompting two widely used, open-source T2I diffusion-based models: Stable Diffusion 2.1 \cite{rombach2022highresolutionimagesynthesislatent} and Stable Diffusion 3.5 Medium \cite{esser2024scalingrectifiedflowtransformers}, adopting the same protocol. In both datasets, images vary by race\footnote{The term \emph{race} is used as in standard ML datasets, acknowledging it as a social construct distinct from \emph{ethnicity} \cite{APA_Race,APA_Ethnicity}. This study does not seek to promote or reify racial categories in AI, but rather to critically examine how AI systems encode and propagate biases linked to socially constructed categories such as race \cite{doh2025position}.} (Asian, Black, White), gender (woman, man), and five attribute pairs associated with lookism in humans (attractiveness halo effect) \cite{gulati2024beautifulgoodattractivenesshalo}: attractive vs.\ unattractive, intelligent vs.\ unintelligent, trustworthy vs.\ untrustworthy, sociable vs.\ unsociable, and happy vs.\ unhappy. Prompts followed the format:
< \textit{Front photo of a [attribute] [race] [gender]} >
with 200 images generated per attribute–race–gender triplet. Additionally, a ``Neutral'' set (200 images per race–gender combination) was generated without any attribute descriptors in the prompt, to serve as a baseline to measure each of the model's default tendencies.

To isolate facial features from potential confounding factors, all generated images were cropped to focus exclusively on the facial region, removing clothing, backgrounds, and other contextual elements that could introduce additional biases into the analysis. This pre-processing ensures that all subsequent analyses examine aesthetic discrimination based solely on facial characteristics rather than external visual cues. 

Note that in this research we neither define nor measure facial attractiveness, 
but focus instead on analyzing how T2I models associate attractiveness, or its lack thereof, with other positive and negative attributes.

\subsection{Research Questions}
\label{sec:rq_main}
The main research question addressed in this paper is as follows:

\textbf{(RQ) How does algorithmic lookism operate as a systematic form of aesthetic discrimination across AI vision systems, from generation to classification, and what systematic patterns of harm does it produce?}

This overarching question is articulated through three research questions, each addressed with specific methodological approaches, as illustrated in Figure \ref{fig:exsd35} :

\textbf{(RQ1) Do synthetic facial images generated by diffusion models exhibit algorithmic lookism, (\emph{i.e.}, an implicit correlation between attractiveness and unrelated positive attributes) and how do such associations vary across demographic groups?}

To tackle this research question and drawing from the literature of this cognitive bias in humans \cite{gulati2024beautifulgoodattractivenesshalo}, we study the existence of a systematic association between four socially desirable attributes (happiness, sociability, trustworthiness and intelligence) and the model’s operationalization of facial attractiveness (RQ1.1). We also assess whether the strength of this correlation can be measured quantitatively and study it on different demographic groups (RQ1.2).

\paragraph{(RQ1.1) Algorithmic Lookism Reference Point Analysis}
This analysis addresses our primary research question by examining whether T2I models systematically associate positive attributes with attractiveness and negative attributes with unattractiveness. We developed a distributional approach that considers all pairwise distances between samples to capture the complete distributional characteristics of attribute associations. We define the following sets:

\begin{tabular}{ll}
Gender (\(G\)) & \{man, woman\} \\
Race (\(R\)) & \{White, Black, Asian\} \\
Attributes (\(A\)) &
\begin{tabular}[t]{@{}l@{}}
\{happy/unhappy; intelligent/unintelligent; neutral; 
sociable/unsociable; \\ trustworthy/untrustworthy; 
 \}
\end{tabular}
\end{tabular}

Note that, although \textit{attractive} and \textit{unattractive} were generated in the same way as the other attributes, they are not included in \(A\) above since they serve as the reference categories in the distance calculations described below.

Each \textbf{demographic group} is defined as a pair \( g = (\text{gender}, \text{race}) \in G \times R \).  Thus, for each demographic group \( g = (\text{gender}, \text{race}) \) and attribute \( a \in A \), we compute two distance distributions:

\begin{align}
\mathcal{D}_g^{(a \rightarrow \text{attractive})} &= \left\{\|\mathbf{e}_i^{(a)} - \mathbf{e}_j^{(\text{attractive})}\|_2 : i = 1,\ldots,N_a, j = 1,\ldots,N_{\text{att}}\right\} \\
\mathcal{D}_g^{(a \rightarrow \text{unattractive})} &= \left\{\|\mathbf{e}_i^{(a)} - \mathbf{e}_k^{(\text{unattractive})}\|_2 : i = 1,\ldots,N_a, k = 1,\ldots,N_{\text{unatt}}\right\}
\end{align}

where $\mathbf{e}_i^{(a)}$ represents the $i$-th embedding vector for attribute $a$ in a given embedding space, and $N_a$, $N_{\text{att}}$, $N_{\text{unatt}}$ denote the number of samples for attribute $a$, attractive, and unattractive categories, respectively.

We operationalize algorithmic lookism via a distributional measure $L_g^{(a)}$, defined as the difference between the expected values of the two distance distributions:

\begin{equation}
L_g^{(a)} = \mathbb{E}[\mathcal{D}_g^{(a \rightarrow \text{unattractive})}] - \mathbb{E}[\mathcal{D}_g^{(a \rightarrow \text{attractive})}]
\end{equation}

A positive value of $L_g^{(a)}$ indicates that attribute $a$ is systematically closer to ``attractive'' samples, while a negative value $L_g^{(a)} < 0$ suggests systematic proximity to ``unattractive'' samples. Therefore, under the algorithmic lookism hypothesis, positive/negative attributes are expected to yield positive/negative values of $L_g^{(a)}$, respectively.

For each attribute–demographic group combination, we perform statistical comparisons between the two distance distributions. We first assess distributional assumptions using the Shapiro–Wilk test for normality and Levene's test for homogeneity of variances, then apply the most appropriate statistical test (independent t-test, Welch's t-test, or Mann–Whitney U test). Effect sizes are quantified using Cohen's $d$ to assess the practical significance of observed differences.

\paragraph{(RQ1.2) Cross-Attribute Correlation Analysis}

This analysis examines how strongly attractiveness and positive attributes are correlated at the representational level in the learned embedding space. 

For each attribute with positive/negative polarity (\emph{i.e.}, all attributes but ``neutral''), let $a^+$ and $a^-$ denote the positive (\emph{e.g.}, trustworthy) and negative (\emph{e.g.}, untrustworthy) version of the attribute, respectively. 

We define the Correlation Lookism Strength (C) per demographic group pair $g = (\text{gender}, \text{race}) \in G \times R $, as:

\begin{equation}
C^{(g)} = \mathbb{E}[\rho(\mathbf{e}_i^{(\text{attractive})}, \mathbf{e}_j^{(a^+)})] - \mathbb{E}[\rho(\mathbf{e}_i^{(\text{attractive})}, \mathbf{e}_k^{(a^-)})]
\end{equation}

where $\rho(\mathbf{e}_i^{(\text{attractive})}, \mathbf{e}_j^{(a^+)})$ represents the correlations between the individual embedding vectors of attractive individuals and the embedding vectors of positive attributes, and analogously for negative attributes.

We compute all pairwise correlations between individual attractive samples and all samples created with positive/negative attributes, then test whether the mean correlation difference is significantly greater than zero using the Mann-Whitney U test. Under the algorithmic lookism hypothesis, we expect $ C^{(g)} > 0$ across demographic groups, indicating stronger correlations between attractive and positive attributes than between attractive and negative attributes.

To quantify the relationship between attractiveness and other attributes, we extracted features from the generated images using two complementary embedding spaces: CLIP \cite{radford2021learningtransferablevisualmodels}, which provides general-purpose visual–semantic representations learned from diverse image–text pairs, and ArcFace \cite{serengil2021lightface,deng2019arcface}, which offers face-specific embeddings trained on facial recognition tasks. This combination ensures that our findings are not dependent on a single representational framework, leveraging both general contextual cues and specialized facial features. By using both, we can validate the robustness of our bias measurements across different embedding spaces and mitigate concerns that our results might be artifacts of a particular learned representation.

\textbf{(RQ2) Does algorithmic lookism impact the performance of downstream tasks, particularly gender classification?}

To address RQ2, we investigate whether algorithmic lookism in T2I models 
propagates to downstream classification tasks. Specifically, we examine whether gender classifiers exhibit differential performance on faces generated with positive versus negative attribute prompts, and whether systematic misclassification patterns emerge based on the attributes used in generation.

We evaluate gender classification performance using three widely-adopted face analysis models: InsightFace \cite{ren2023pbidr}, DeepFace \cite{serengil2021lightface}, and FairFace \cite{karkkainenfairface}. For each demographic group pair $g = (\text{gender}, \text{race}) \in G \times R $ and attribute $a \in A$, we compute the gender classification accuracy as:

\begin{equation}
\text{Accuracy}_{g,a} = \frac{1}{N_{g,a}} \sum_{i=1}^{N_{g,a}} \mathbb{I}[\hat{y}_i = y_i]\times 100
\end{equation}

where $N_{g,a}$ is the total number of images for demographic group $g$ and attribute $a$, $\hat{y}_i$ is the predicted gender label for image $i$, $y_i$ is the true gender label specified in the generation prompt, and $\mathbb{I}[\cdot]$ denotes the indicator function. A classification is considered correct when the predicted gender matches the gender specified in the generation prompt.

We compare the classification performance across attributes to assess whether the faces generated with positive versus negative attributes exhibit systematic differences in gender classification accuracy. Misclassification rates are analyzed across all gender–race combinations to identify biases that disproportionately affect certain demographic groups based on the aesthetic encoding in generated faces.

\textbf{(RQ3) What aesthetic patterns and representational constraints emerge in AI-generated faces beyond quantifiable lookism metrics?}

Embedding-based analyses reveal systematic lookism but cannot capture all dimensions of aesthetic constraint. We conduct a qualitative visual analysis to document how representational narrowing operates through age homogenization, gendered exposure patterns, and geographic reductionism across SD 2.1 and 3.5.

\section{Results}
We report the results of investigating how \textit{algorithmic lookism} operates across three levels: encoding attractiveness-positive attribute associations in generation (RQ1); propagating to classification and creating gendered disparities (RQ2); and intensifying through aesthetic narrowing in newer models (RQ3).

\subsection{\textit{(RQ1) Do synthetic facial images generated by diffusion models exhibit algorithmic lookism?}}
\label{sec:rq1}
\paragraph{(RQ1.1) Algorithmic Lookism Reference Point Analysis}
\label{sec:rq1.1}

\begin{figure}
    \centering
    \includegraphics[width=\linewidth]{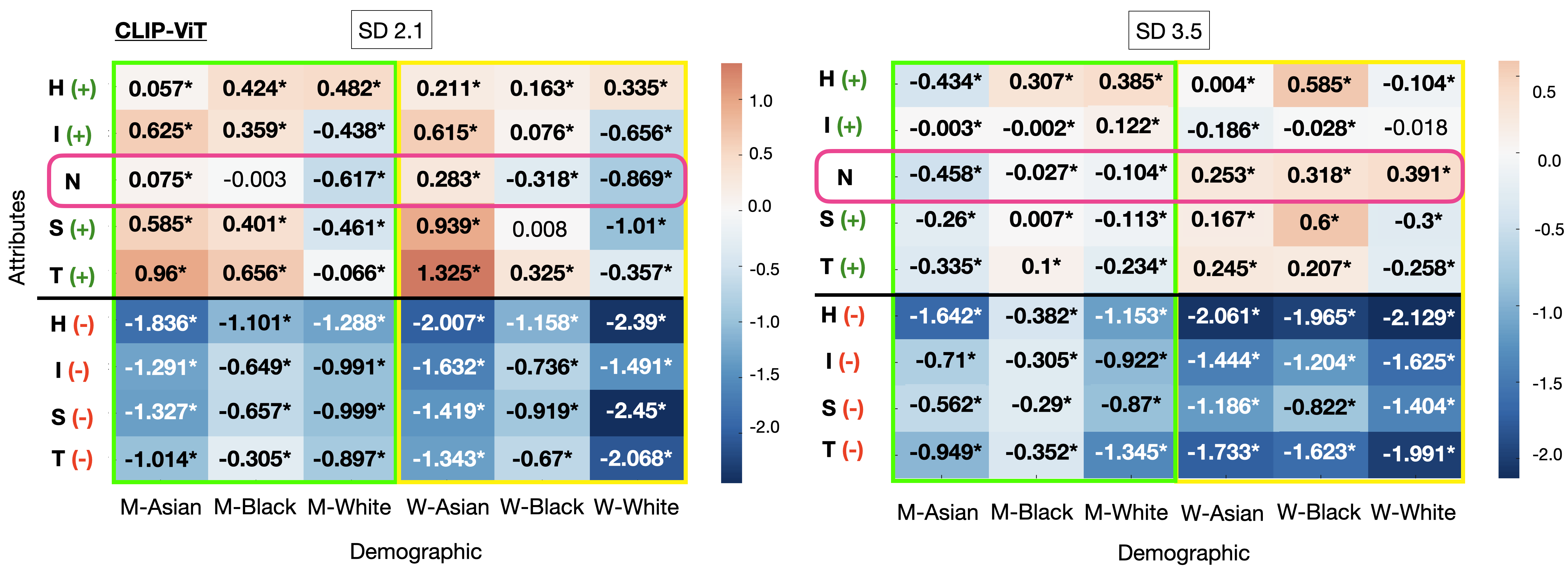}
    \caption{Distributional measure scores ($L_g^{(a)}$) across demographic groups in images generated with Stable Diffusion 2.1 (left) and 3.5 (right), using CLIP embeddings. Negative values (blue) indicate closeness to unattractive faces; positive values (red) to attractive ones. An asterisk (*) indicates statistical significance ($p<0.05$). The neutral trait is highlighted with a pink box. The results for men/women are highlighted with a green/yellow box, respectively. A = Attractiveness, H = Happiness, I = Intelligence, S = Sociability, T = Trustworthiness, N = Neutral.}

    \label{fig:lookism_comparison_clip}
\end{figure}

\begin{figure}
    \centering
    \includegraphics[width=\linewidth]{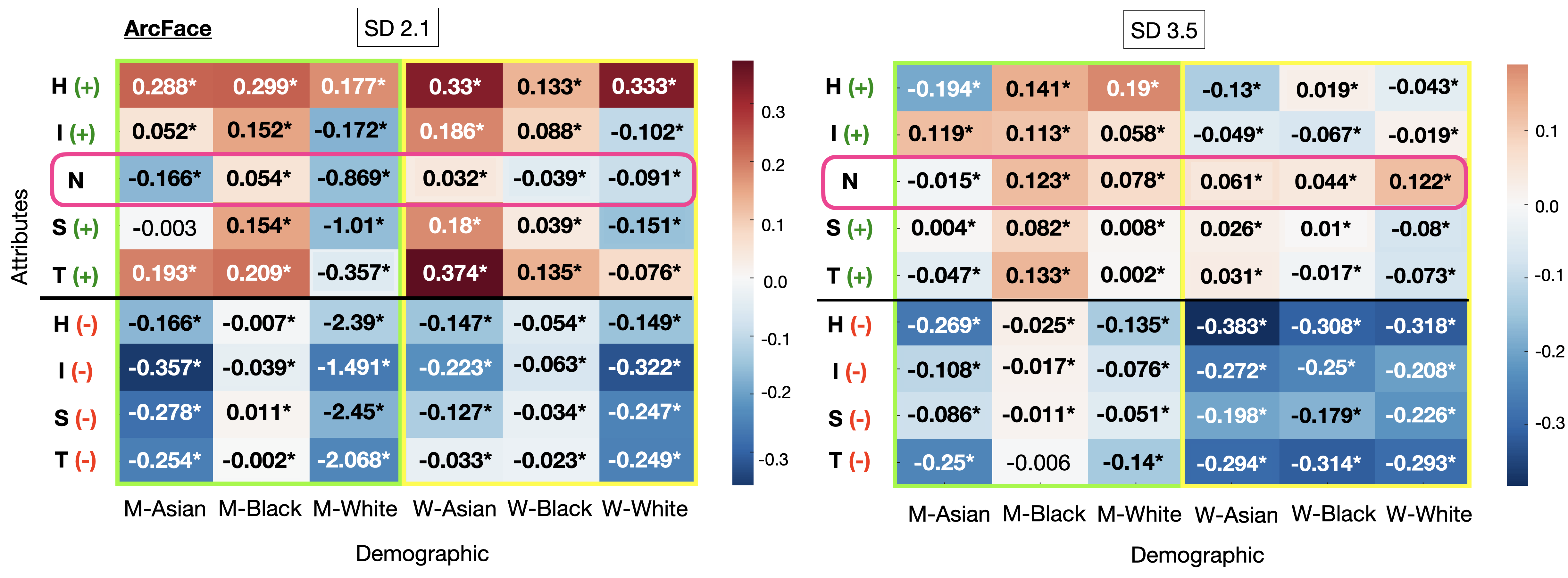}
    \caption{Distributional measure scores ($L_g^{(a)}$) scores across demographic groups in images generated with Stable Diffusion 2.1 (left) and 3.5 (right), using ArcFace embeddings. Negative values (blue) indicate closeness to unattractive faces; positive values (red) to attractive ones. An asterisk (*) indicates statistical significance ($p<0.05$). The neutral trait is highlighted with a pink box. The results for men/women are highlighted with a green/yellow box, respectively. A = Attractiveness, H = Happiness, I = Intelligence, S = Sociability, T = Trustworthiness, N = Neutral.}

    \label{fig:lookism_comparison_arcface}
\end{figure}

Figures \ref{fig:lookism_comparison_clip} and \ref{fig:lookism_comparison_arcface} depict the distributional measure scores across demographic groups in the synthetic faces. As seen in the figures, both SD 2.1 and 3.5 exhibit systematic algorithmic lookism: faces generated with positive attributes cluster near ``attractive'' references, while negative attributes align with ``unattractive'' (reflected as positive/negative values of $L_g^{(a)}$, respectively) validated both on the ArcFace and CLIP embeddings and all demographic groups, with stronger effects in CLIP. All values except for one--corresponding to sociable Asian men--are statistically significant.

Critically, ``neutral'' faces (outlined in pink in the Figures) reveal gendered aesthetic defaults. In SD 2.1, neutral faces skew toward unattractiveness with limited demographic variation. SD 3.5 reverses this: neutral faces approach attractiveness, but only for women. In the CLIP embedding space, neutral female faces are closer to attractive than unattractive faces and vice-versa for male faces, revealing a structurally encoded default beautification bias for women that operates even in the absence of explicit prompts. This growing gender asymmetry shows femininity increasingly subject to aesthetic optimization, embedding normative assumptions that link female identity to beauty.

These findings demonstrate that structural correlations between attractiveness and unrelated traits are embedded in the generative space, supporting the algorithmic lookism hypothesis. Gender- and race-specific patterns, along with an analysis of neutral defaults, are detailed in Section~\ref{sec:rq1.2}.


\paragraph{(RQ1.2) Cross-Trait Correlation Analysis and Neutral Defaults}
\label{sec:rq1.2}

The demographic breakdown shows that algorithmic lookism affects gender and racial categories to different degrees, with some groups experiencing stronger and more consistent aesthetic bias than others. We compute the Correlation Lookism Strength $C^{(g)}$  defined in Section \ref{sec:rq_main} to analyze how attractiveness correlates with positive versus negative traits across demographic groups. Statistical significance is assessed using the Mann-Whitney U test. For cross-trait correlations, unless otherwise noted, reported differences are statistically significant with $p < 10^{-16}$.

In the case of SD~2.1, CLIP embeddings show the strongest correlations for Asian women ($C^{(g)} = +0.084$), followed by Asian men ($C^{(g)} = +0.060$) and Black women ($C^{(g)} = +0.053$). The effect is weaker for White subjects: White women ($C^{(g)} = +0.013$) and White men ($C^{(g)} = +0.023$), indicating reduced coupling between attractiveness and positive attributes for these groups. ArcFace embeddings confirm the same demographic ordering: Asian women ($C^{(g)} = +0.099$), Asian men ($C^{(g)} = +0.073$), with all groups statistically significant.

Neutral faces reveal the models’ aesthetic defaults. In the CLIP embedding space, only the faces of Asian women exhibit significant displacement toward attractiveness ($C_{\text{neutral}}^{(g)} = +0.009$, $p < 10^{-16}$), while all other groups show non-significant effects ($p > 0.05$), with a slight shift toward unattractiveness (negative $C$ values ranging from $-0.003$ to $-0.068$). 

The results in the ArcFace embedding space show a similar pattern: all groups are weakly shifted, on average, toward unattractiveness (negative $C$ values ranging from $-0.009$ to $-0.088$), but none reaches statistical significance ($p > 0.05$ for all groups). These results indicate that aesthetic bias in SD~2.1 operates primarily through explicit trait conditioning rather than through systematic neutral defaults.

Regarding SD~3.5, in the CLIP embedding space, the strongest effects appear for White men ($C^{(g)} = +0.065$) and women across racial groups (Black: $C^{(g)} = +0.061$; Asian: $C^{(g)} = +0.052$; White: $C^{(g)} = +0.051$), all $p < 10^{-16}$. In the ArcFace embeddings, the faces of women exhibit stronger correlations between attractiveness and positive traits (White: $C^{(g)} = +0.056$; Black: $C^{(g)} = +0.055$; Asian: $C^{(g)} = +0.036$), with Asian men showing no significant effect ($C^{(g)} = -0.006$, $p > 0.05$).

Neutral defaults in SD~3.5 reveal pronounced gendered patterns with significantly stronger effects than SD~2.1. In the CLIP embedding space, all female faces are shifted toward attractiveness (White: $C_{\text{neutral}}^{(g)} = +0.021$; Black: $C_{\text{neutral}}^{(g)} = +0.006$; Asian: $C_{\text{neutral}}^{(g)} = +0.003$, all $p < 10^{-16}$), while male faces show non-significant shift toward attractiveness (negative $C$ values, $p > 0.05$). In the ArcFace embedding space, the pattern intensifies: most groups are strongly shifted toward attractiveness (White women: $C_{\text{neutral}}^{(g)} = +0.073$; Asian women: $C_{\text{neutral}}^{(g)} = +0.068$; Black women: $C_{\text{neutral}}^{(g)} = +0.049$; White men: $C_{\text{neutral}}^{(g)} = +0.014$; Black men: $C_{\text{neutral}}^{(g)} = +0.012$, all $p < 10^{-16}$), with Asian men as the only exception showing displacement toward unattractiveness ($C_{\text{neutral}}^{(g)} = -0.011$, $p > 0.05$).

These findings demonstrate that algorithmic lookism operates through both explicit trait associations and implicit aesthetic defaults, with strength and demographic patterns varying across model versions. In SD~2.1, racialized women, particularly Asian and Black women, exhibit the strongest cross-trait correlations, while neutral defaults remain weak and non-significant. In SD~3.5, the gender gap intensifies: women show stronger cross-trait correlations (mean $C^{(g)} = 0.049$--$0.055$) than men (mean $C^{(g)} = 0.019$--$0.033$) across racial groups, and neutral defaults become systematic, with all female faces significantly aligned with attractive references ($p < 10^{-16}$) while male faces show non-significant patterns in the CLIP embedding space.

Overall, the results in Sections \ref{sec:rq1.1} and \ref{sec:rq1.2} confirm that positive attributes are more strongly correlated with facial attractiveness than negative attributes across all demographic groups, with substantially stronger effects for women. These results indicate that aesthetic bias is not only expressed through explicit trait conditioning but is also increasingly embedded in the models' default generative representations, reflecting a demographic asymmetry in how social value is visually encoded.

\subsection{\textit{(RQ2) Does algorithmic lookism impact the performance of gender classification?}}
\label{sec:rqé}

\label{sec:classification}

\begin{figure}
    \centering
    \includegraphics[width=\linewidth]{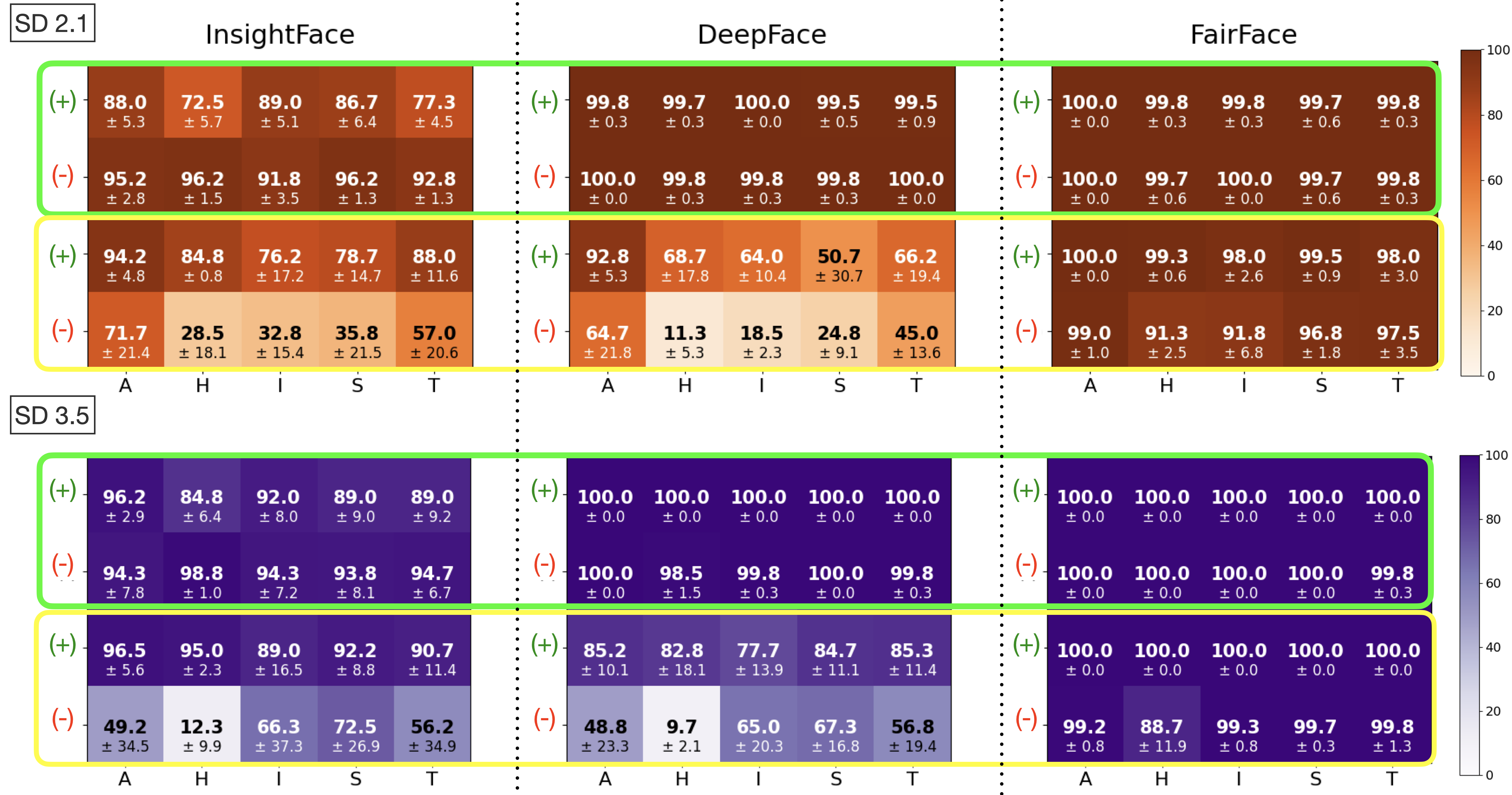}
    \caption{Heatmaps of gender classification accuracy (Mean ± Std) for InsightFace, DeepFace, and FairFace for \textbf{SD 2.1} (orange) and \textbf{SD 3.5} (purple) . A = Attractiveness, H = Happiness, I = Intelligence, S = Sociability, T = Trustworthiness. Performance corresponding to female (Women)/male (Men) faces is highlighed in Yellow \colorsquare{c_female} / Green \colorsquare{c_male}, respectively. \underline{Accuracies in classifying neutral faces \textbf{SD 2.1}} (\%):   InsightFace: (W) 80.9 ± 10.7 (M) 87.5 ± 1.5, DeepFace: (W) 51.1 ± 9.0 (M) 99.6 ± 0.1, FairFace: (W) 96.8 ± 1.4 (M) 98.9±0.4. \underline{Accuracies in classifying neutral faces \textbf{SD 3.5}} (\%):    InsightFace: (W) 83.0 ± 17.5 (M) 89.8 ± 5.7, DeepFace: (W) 76.5 ± 14.3 (W) 100.0 ± 0, FairFace: (W) 99.8 ± 0.2 (M) 100 ± 0. }

    \label{fig:classification21}
\end{figure}

Figure \ref{fig:classification21} reports classification performance by model and gender, with race categories aggregated for ease of interpretation in the main text. The disaggregated results by race are provided in Appendix \ref{appendix:classification}.
As shown, performance varies significantly depending on the attributes and the gender of the generated image.

All three classifiers achieve stable, high accuracy on male faces across attributes and models: DeepFace $\geq$ 98.5\%, FairFace 99.7-100\%, InsightFace 72.5-98.8\%. This stability persists across racial categories, confirming algorithmic bias disproportionately affects female classification regardless of identity or model version (Appendix \ref{appendix:classification}).

Conversely, the classification performance of the three algorithms on female faces demonstrates considerably different patterns, with large variations depending on the attributes and exhibiting a systematic deterioration in specific model-attribute combinations on the two generated datasets. On SD 2.1-generated images, the performance of the classification algorithms on female faces with negative attributes shows a substantial drop: the classification accuracy on ``unhappy'' female faces is only 28.5\% for InsightFace and 11.3\% for DeepFace, while the classification accuracy of ``happy'' female faces reaches 84.8\% and 68.7\%, respectively. Intelligence shows similar patterns, with a classification performance on faces of ``unintelligent'' women of only 32.8\% (InsightFace) and 18.5\% (DeepFace), compared to 76.2\% and 64.0\% for their ``intelligent'' counterparts. FairFace maintains more stable performance across different attributes of female faces but still exhibits systematic sensitivity to the attributes, with accuracy ranging from 91.3\% (for unhappy female faces) to 100\% (for attractive female faces). 

Classification performance on SD 3.5-generated images reveals a paradoxical pattern of simultaneous improvement and deterioration, depending on the model and the attribute. FairFace achieves near-perfect performance on faces created with positive attributes (100\%) and substantial improvements on faces of women corresponding to negative attributes (88.7\%-99.7\%). DeepFace yields mixed results, with some improvements on female faces created with certain negative attributes, like ``unintelligence'' (65.0\% vs 18.5\% on SD 2.1-generated images), but persistent poor performance on female faces with other attributes, such as ``unhappiness'' (9.7\%). 

Most concerning is InsightFace's dramatic deterioration of performance on SD 3.5: classification accuracy on ``unhappy'' and ``unattractive'' female faces collapses to 12.3\% and 49.2\%, respectively.
This significant misrecognition rates reveal compounding discrimination: women are already misclassified at higher rates than men on neutral faces, and the disparities intensify substantially for faces generated with negative attributes. Aesthetic non-conformity compounds gendered vulnerability.

A disaggregated analysis by racial category (detailed in Appendix \ref{appendix:classification}) reveals that these discriminatory patterns operate through complex intersectional mechanisms on both generated datasets. The drop in performance is the most severe on faces of Black women: InsightFace's performance for ``unhappy" and ``unattractive" Black women drops to a catastrophic 1.0\% on SD 3.5-generated images (from 9.5\% on SD 2.1-generated images) and 10.0\% (from 53.0\%), respectively. In the case of faces of Asian women, we observe a marked yet less severe deterioration of performance. Interestingly, the performance on faces of White women exhibits more variable patterns. 

An analysis of the performance of the gender classification algorithms on neutral faces,  reveals baseline discriminatory patterns that operate even without explicit attribute cues. On SD 2.1-generated neutral images, significant gender gaps in performance emerge: 6.6 points (W: 80.9\% vs M: 87.5\%), 48.5 points (W: 51.1\% vs M: 99.6\%) and 2.1 points (W: 96.8\% vs M: 98.9\%) for InsightFace, DeepFace and FairFace, respectively. The performance on SD 3.5-generated neutral images shows mixed evolution: FairFace approaches near-perfect equity (W: 99.8\% vs M: 100\%), DeepFace maintains a substantial performance gap (W: 76.5\% vs M: 100\%), and InsightFace shows increased variability (W: 83.0 ± 17.5\% vs M: 89.8 ± 5.7). An intersectional analysis reveals that the gender classification performance on faces of Black women is the lowest in both datasets for InsightFace (39.5\% SD 2.1, 58.5\% SD 3.5) and DeepFace (32.5\% SD 2.1, 58.5\% SD 3.5), while FairFace maintains consistently high performance ($\geq$ 95.5\%) across all female racial categories. These neutral baselines demonstrate that discriminatory patterns exist independently of the attributes, providing the foundation upon which attribute-based disparities are amplified.

\begin{figure}[h]
    \centering
    \includegraphics[width=0.8\linewidth]{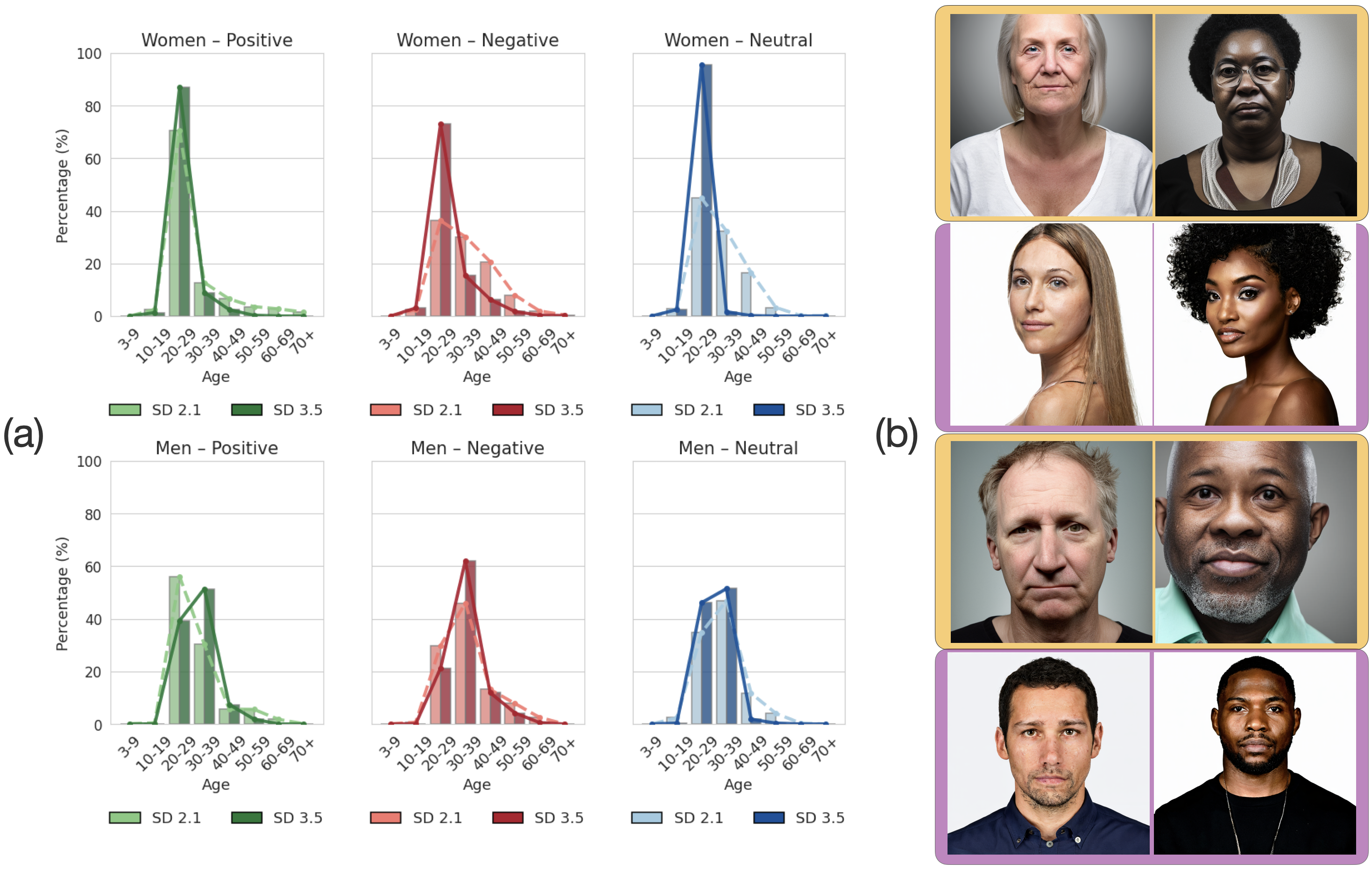}
    \caption{\textbf{(a)} Age homogeneization and agism in the generated images: Age distribution of synthetic faces by gender and associated attributes in SD 2.1 and SD 3.5. The images created with positive attributes tend to depict younger individuals than the images created with negative or neutral attributes. Age classification was performed using FairFace. Top/bottom rows correspond to images of women/men, respectively. \textbf{(b)} Generated faces from SD 2.1 (orange) and SD 3.5 (purple) for the ``Neutral'' prompt for ``White'' and ``Black'' categories, illustrating the shift toward visually younger appearances. }

    \label{fig:age}
\end{figure}

Performance levels below 10\% in attribute-specific conditions reveal ``algorithmic invisibility'', \emph{i.e.}, the systematic failure to recognize the legitimacy of certain intersectional identity presentations. However, even the neutral face analysis reveals baseline discriminatory patterns, with some groups experiencing significantly lower recognition rates that provide the basis for attribute-based amplification. The systematic nature of these disparities reveals how biases create cascading discrimination effects throughout AI systems. The consistency of bias patterns across independently trained classification models when applied to the same generated datasets suggests the operation of a coherent normative infrastructure that determines which identity configurations are considered algorithmically legitimate. This infrastructure operates at multiple levels: neutral faces establish baseline hierarchies of recognition, while the attributes amplify these disparities. In the case of faces of females, positive attributes increase the classification performance and negative attributes decrease it, creating systematic barriers to accurate classification. This behavior is reversed in the case of faces of males, where for some models, the classification performance deteriorates on faces created with positive attributes. 

While embedding-based analyses (RQ1) and classification performance (RQ2) reveal systematic algorithmic lookism through measurable disparities, visual examination uncovers additional aesthetic patterns operating beyond these quantitative measurements. Next we describe a qualitative analysis to document how aesthetic discrimination manifests in age representation, gendered exposure, and geographic diversity.

\subsection{\textit{(RQ3) What patterns emerge in AI-generated faces beyond quantifiable lookism metrics?}}

\label{sec:qualitative_analysis}
\begin{figure}[h]
    \centering
    \includegraphics[width=0.85\linewidth]{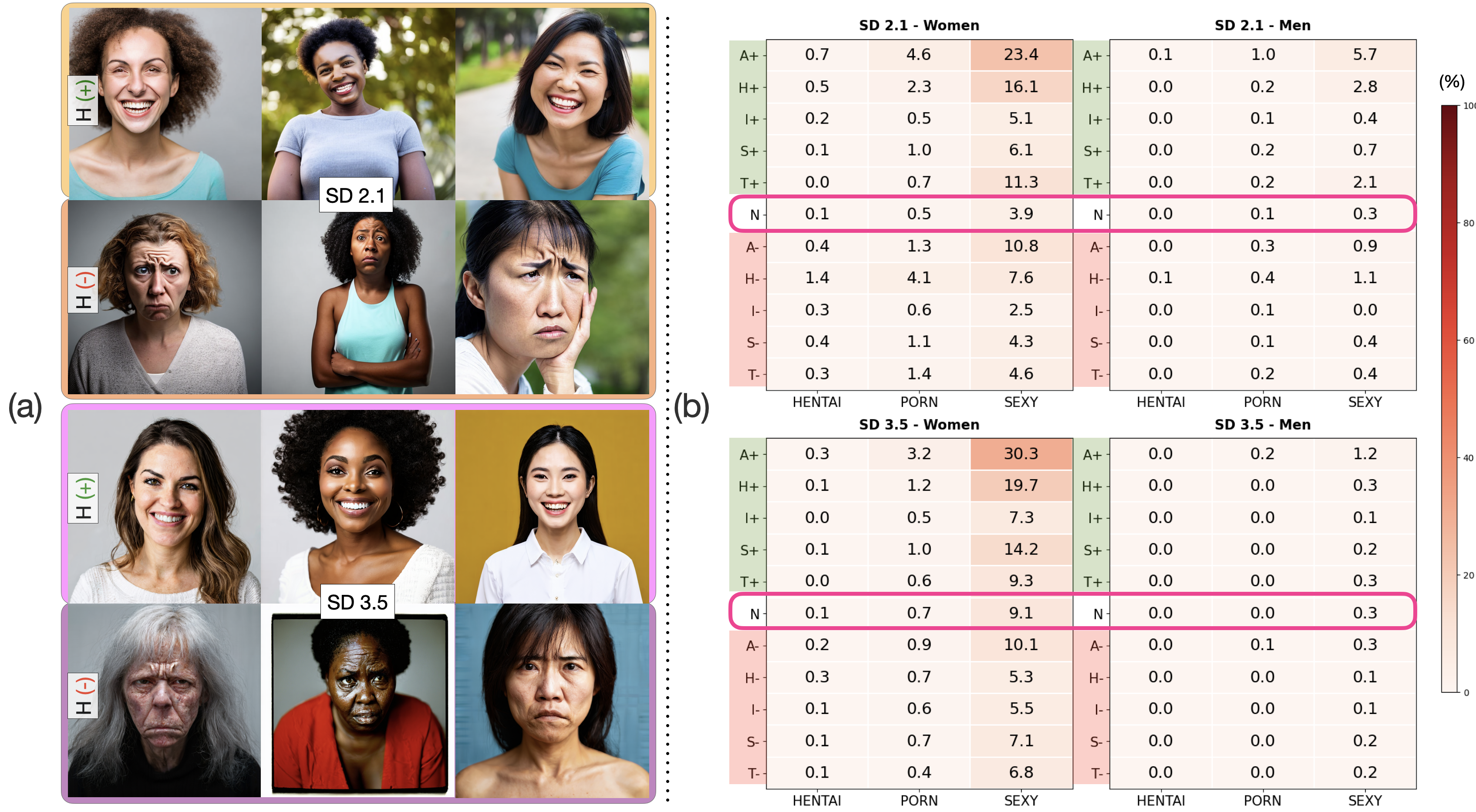}
    \caption{\textbf{(a)} Generated faces of women with the attributes “happy” (H\textcolor{darkgreen}{(+)}) and “unhappy” (H\textcolor{red}{(-)}) across Stable Diffusion 2.1 (orange) and 3.5 (purple). \textbf{(b)} NSFW detection rates (hentai, porn, sexy) by positive (green)/ negative (red) attributes, gender, and 
model version. Neutral baseline results are highlighted in pink.}

    \label{fig:tobehappy}
\end{figure}

Beyond the previously described statistical analyses, we examined the visual content of the generated uncropped images, recognizing that visual representation fundamentally shapes human categorization and judgment. In this section, we report the results of a qualitative assessment of the generated faces. Our analysis reveals systematic patterns that complement the quantitative findings. Most notably, while SD 3.5 produces more convincing and polished images compared to SD 2.1, this apparent technical improvement comes with a troubling trade-off: the newer model demonstrates a pronounced shift toward idealized representations that codify Eurocentric normative aesthetics as the default standard for visibility. This creates a paradox where images become stylistically and technically more realistic while simultaneously shifting towards being less representative of actual human diversity and heterogeneity, hence becoming far more unrealistic in terms of representation. In particular, we observe the following patterns:

\paragraph{(a) The Progressive Age Homogenization and Algorithmic Ageism}
We analyzed the age distribution of the generated faces using FairFace age classification, as shown in Figure \ref{fig:age}. The data confirms a systematic shift toward younger appearances in SD 3.5 compared to SD 2.1 for both men and women. Despite SD 2.1's lower image quality and less photorealistic output, it exhibits greater age diversity across demographic groups. This heterogeneity extends beyond age to include other aesthetic variations, most notably the presence or absence of makeup among women, particularly evident in our neutral baseline generations (Figure \ref{fig:tobehappy}.(a)). Furthermore, we observe that positive attributes generally lead to the creation of images of younger individuals (particularly women) when compared to negative attributes, where there is more diversity in the distribution of ages and the generated faces tend to correspond to older individuals (Figure \ref{fig:age}.(b)).

\paragraph{(b) The Happiness-Beauty Conflation}
One of the most revealing patterns emerges in the representation of attributes, particularly happiness (happy/unhappy). Among all tested attributes, faces created with the ``unhappy" attribute are the most visually caricatured. 
In SD 2.1, unhappiness is primarily conveyed through facial expressions: exaggerated frowns, down-turned mouths, and melancholic eyes. In SD 3.5, however, ``unhappy'' takes on a broader aesthetic meaning: faces are not only sad, but systematically appear older, less groomed, and devoid of makeup—especially among women (see Figure \ref{fig:tobehappy}.(a)). This shift suggests that happiness is no longer represented solely as an emotion, but increasingly stylized as an aesthetic ideal tied to age, fitness, and self-care. In SD 3.5, conceptions of what \emph{happy} means are rendered into looking young, sleek, and polished, and not merely into smiles and joyful facial expressions untied from one's appearance. In this way, the representation of happiness extends far beyond emotional expression, revealing how affective cues are reinterpreted as markers of both self-worth and aesthetic, and hence social, desirability.

\paragraph{(c) Gendered Exposure}
Across both datasets, women are consistently generated with higher frequency of exposed skin, particularly in the neck, shoulders, and cleavage/décolletage areas. This pattern holds regardless of the specified attribute and persists uniformly from SD 2.1 to SD 3.5, indicating that skin exposure has become a recurring feature of how these models represent femininity. This observation is further supported by the analysis of NSFW content reported in Figure \ref{fig:tobehappy}.(b), which shows that images of women are flagged as NSFW at higher rates than their male counterparts (Further details in Appendix \ref{app:nsfw_analysis}).

\paragraph{(d) Geographic Reductionism in the Representation of ``Asian" Faces}
The treatment of the ``Asian" category exemplifies a form of geographic reductionism that reveals the Western-centric assumptions embedded in these models. Both SD 2.1 and SD 3.5 reproduce exclusively Northeastern Asian traits, essentially corresponding to Chinese, Korean, and Japanese phenotypes, while completely neglecting the vast diversity of the Asian continent. Here, a centralized vision conflates an entire continent with specific regional appearances that reflect hegemonic visual priors likely shaped by Western views, effectively invisibilizing South Asian, Southeast Asian, Central Asian, and Middle Eastern populations. 

\section{Discussion: Convergent Mechanisms of Algorithmic Harm}  
From our evaluation, we conclude that both generative and discriminative computer vision systems exhibit \textit{algorithmic lookism}. Furthermore, our findings show that the algorithmic lookism operates not as an isolated glitch but as a systemic mechanism that discursively shapes both digital representation and computational recognition. This bias reflects and reinforces broader structures of inequality, particularly those shaped by neoliberal rationality, racialized and gendered norms, and platform-driven optimization. Below, we discuss four interrelated implications of our work.

\paragraph{Neoliberal Rationality and the Economics of (In)visibility}
Our findings reveal systematic lookism: positive attributes (intelligent, trustworthy, sociable, happy) align with ``attractive" representations, 
negative with ``unattractive", with stronger effects on images of women and particularly of Black and Asian women.

This aesthetic hierarchy cannot be understood outside neoliberal rationality, which, following Wendy Brown \cite{brown2003neo,brown2015undoing}, operates as a form of \textit{governmentality} \footnote{The term governmentality, coined by Michel Foucault in his 1978–79 lectures, refers to “the ensemble formed by institutions, procedures, analyses and reflections, calculations, and tactics” that enable the exercise of power over populations and the subject—where political economy becomes the dominant form of knowledge (Foucault, 1978–79). Derived from the French \textit{gouverner} and \textit{mentalité}, the term underscores how modes of thought structure governance through “the conduct of conduct”, encompassing practices that range from self-governance to the governance of others.} that not only evaluates, but constitutes and governs subjects through logics of market-driven productivity, appeal, and self-optimization. Here, marketability defines what Sarah Banet-Weiser calls ``economies of visibility" that  ``increasingly structure not just our mediascapes, but our cultural and economic practices and daily lives" \cite{banet2015keynote}. Under this framework, facial appearance becomes a proxy for individual and social value defined through market demands and capitalization, while visibility itself becomes a commodity. What cannot be optimized aesthetically is gradually rendered invisible, not through deletion, but through erasure by omission, under-representation, and mis-recognition.

\paragraph{From Representation to Recognition: Aesthetic Filtering as a Structural Barrier}
Aesthetic hierarchies propagate from generation to classification: female faces generated with negative attributes suffer systematic mis-classification, revealing how aesthetic norms govern not only depiction but computational recognition. Faces deviating from normative ideals become algorithmically illegible—less \textit{desirable}, the more likely misread or erased. This compounds representational harm with computational exclusion. As Banet-Weiser et al. \cite{banet2020postfeminism} argue, neoliberal visibility requires being seen \textit{correctly} through dominant codes. Generative systems encode these norms; classifiers inherit and enforce them.

\paragraph{Aesthetic Entrepreneurship and the Privatization of Harm}

The evolution from SD 2.1 to SD 3.5 exemplifies how technical advancement intensifies aesthetic constraints. SD 3.5 generates higher quality images, and some metrics suggest progress: certain embedding-space correlation shifts (RQ1), gender classification accuracy improvements (RQ2), and greater photorealism. However, our qualitative analysis (RQ3) also reveals that this apparent improvement masks aesthetic narrowing: the images generated with SD 3.5 are more homogeneous regarding age, with a default beautification of women, and reduced phenotypic variation which is less severe in the images created with SD 2.1. Enhanced curation filters aesthetic non-conformity under the guise of quality control. What appears as technical refinement tightens representational boundaries, where algorithmic legibility requires conformity to narrow aesthetic standards optimized for profitable ideals.
This dynamic resonates with work by Elias \emph{et al.} \cite{elias2017aesthetic}  on aesthetic labor and its logic of gendered entrepreneurship under which women are encouraged to treat their bodies as ongoing projects of self-transformation and branding. In generative AI contexts, this imperative becomes automated and systems reward conformity to learned attractiveness norms while penalizing deviation. Corporate platforms frame outputs as neutral while those falling outside algorithmic standards become anomalous, less recognizable, misclassified, or invisible. Meanwhile, platforms profit from unpaid data labor and attention economies \cite{terranova2022after}, transforming aesthetic compliance into computational capital. 
As Gill \cite{gill2007gender} argues, apparently progressive media environments often reproduce (and profit from) gendered hierarchies through logics of sexualization, desirability, and objectification. AI systems inherit these logics, encoding them into visual defaults that dictate not only what can be seen, but how it must appear to be recognized correctly. Algorithmic lookism does not reflect a system failure, but the successful reproduction of dominant ideologies through computer vision, creating infrastructural harm built into how platforms govern digital aesthetics and mediate social legibility.

\paragraph{Orientalist Regimes and Geographic Reductionism}
Our analysis reveals that both generative models reduce the representation of ``Asian" faces to exclusively Northeastern Asian features (Chinese, Korean, and Japanese phenotypes) while completely failing to represent South Asian, Southeast Asian, Central Asian, and Middle Eastern populations. This can be understood as a computationally encoded, aestheticized iteration of a visually reproduced orientalist logic \cite{said1978orientalism}\footnote{Through his conception of \textit{Orientalism}, Edward Said outlined how the West has historically constructed ideals of the ``Orient" as a homogeneous, exotic, and aestheticized ``other" under terms it could recognize and control. This also materialized through the distortion and erasure of complexity of the Asian continent. SD 2.1 and 3.5 in this sense, replicate the epistemological logic of Orientalism through selective visibility, stylization, and erasure.} rendered through algorithmic selectivity rather than textual discourse, flattening and erasing complexity while reinforcing hierarchies of legibility aligned with Western, neoliberal demands.
This geographic reductionism reflects what Quadri \emph{et al.} \cite{qadri2023ai} term algorithmic Orientalism \cite{said1978orientalism}. Drawing on \cite{breckenridge1993orientalism}, they show how AI systems perpetuate colonial stereotypes that flatten diverse regions into simplified, Western-legible categories. The ``Orient" becomes a singular representation that obscures vast cultural and geographic diversity.
Our findings demonstrate the same dynamic: ``Asian" identity gets conflated with East Asian appearance, invisibilizing entire populations across the continent. This represents another form of systematic exclusion embedded in generative AI—one that transforms historical colonial ways of seeing into computational defaults that determine which communities become visible and recognized.

\paragraph{Limitations}
Several limitations must be acknowledged. Our binary framing of “attractive/unattractive” presumes shared beauty standards without human validation, reducing a culturally variable concept to fixed categories. Similarly, our demographic scope with three racial categories and binary gender excludes mixed-race, non-binary, and other marginalized identities whose experiences with algorithmic lookism may diverge from our findings. Finally, our qualitative study relies on the automatic classification of age which could be biased.

\section{Conclusion}

Through empirical analysis of 26,400 synthetically generated faces, we demonstrate that algorithmic lookism operates as systematic infrastructure in text-to-image generation and gender classification. Stable Diffusion models consistently map positive traits ("happy", "intelligent", "sociable", "trustworthy") onto normatively attractive faces while aligning negative traits with less attractive ones. These patterns disproportionately affect 
women, particularly Black and Asian women, who face both stronger 
aestheticization in generation and higher misclassification rates in downstream systems. The convergence of bias across generation and recognition demonstrates that algorithmic lookism functions as coherent normative framework rather than isolated technical failure.

Comparison between SD 2.1 and 3.5 reveals that while some quantitative metrics suggest reduced bias, qualitative analysis shows enhanced data curation is accompanied by intensified aesthetic constraints—age homogenization, default beautification of women, and reduced phenotypic variation.

Following Hall \cite{hall1997representation}, we understand these patterns as systems of representation shaped by cultural and ideological forces that determine which bodies become visible and how. Generative AI encodes these normative visual codes, making algorithmic lookism a mechanism governing legibility, recognition, and social worth. As Birhane \cite{birhane2021multimodaldatasetsmisogynypornography} warns, multimodal systems risk entrenching stereotypes when visual meaning is inferred from culturally loaded prompts, and "realism" becomes proxy for conformity to dominant norms.

What presents as aesthetic optimization operates as systematic exclusion.
Social norms become computational defaults; divergence is penalized through invisibility or misrecognition. Algorithmic lookism moves from representational concern to infrastructural barrier determining participation in automated societies.

\section{Acknowledgements}
M.D. acknowledges support from the ARIAC project (No. 2010235), funded by the Service Public de Wallonie (SPW Recherche), and funding from the FNRS (National Fund for Scientific Research) for her visiting research at the ELLIS Alicante Foundation. A.G. and N.O. are partially supported by a nominal grant received at the ELLIS Unit Alicante Foundation from the Regional Government of Valencia in Spain (Resolución de la Conselleria de Innovación, Industria, Comercio y Turismo, Dirección General de Innovación), along with grants from the European Union’s Horizon Europe research and innovation programme (ELIAS; grant agreement 101120237) and Intel. A.G. is additionally partially supported by a grant from the Banc Sabadell Foundation. C.C. is supported by the Weizenbaum Institute, whose core funding is provided by the German Federal Ministry of Education and Research (BMBF). Views and opinions expressed are those of the author(s) only and do not necessarily reflect those of the European Union or the European Health and Digital Executive Agency (HaDEA).  


\bibliographystyle{ACM-Reference-Format}
\bibliography{sample-base}

\appendix
\section{Examples of Generated Faces}
\label{appendix:ex_images}
\begin{figure}[h]
    \centering
    \includegraphics[width=\linewidth]{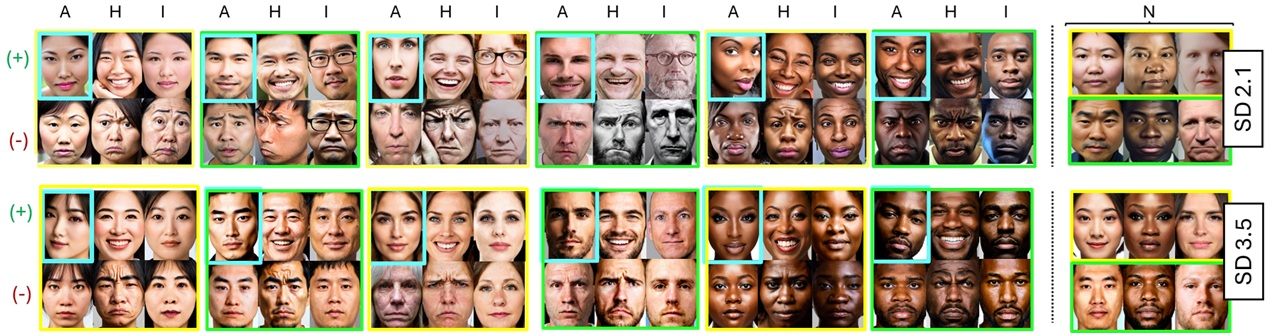}
    \caption{Examples of the generated faces with Stable Diffusion 2.1  and Stable Diffusion 3.5 Medium with positive (+) and negative (-) variations for three attributes (A = Attractiveness, H = Happiness, and I = Intelligence) together with the neutral faces (N = Neutral). Yellow (\colorsquare{c_female}) and green (\colorsquare{c_male}) correspond to images of females and males, respectively. Light Blue (\colorsquare{c3}) borders highlight the faces corresponding to the positive Attractiveness trait.}

    \label{fig:generared_faces}
\end{figure}
\section{Training Differences Between Stable Diffusion 2.1 and 3.5 Medium}
\label{appendix:training_differencs}

While both image sets in our study were generated using the same prompting protocol, the underlying diffusion models differ substantially in architecture, scale, and training pipeline. Stable Diffusion 2.1 \cite{rombach2022highresolutionimagesynthesislatent} was trained on filtered subsets of the LAION-5B dataset \cite{schuhmann2022laion5b}, a large-scale corpus of image-text pairs scraped from the public web. The dataset was preprocessed using automated filters to remove content deemed unsafe or of aesthetically low-quality. Nonetheless, prior studies have pointed out that large-scale web-scraped datasets like LAION-5B often contain pervasive misogyny, explicit content, and stereotypical representations, which can propagate harmful associations along lines of gender, race, and perceived physical appearance \cite{sesha2023bias,birhane2021multimodaldatasetsmisogynypornography,birhane2023into}.

The training data used for Stable Diffusion 3.5 Medium, by contrast, has not been publicly disclosed. However, available documentation highlights significant changes in data curation practices. These include the application of NSFW classifiers, aesthetic ranking systems, semantic deduplication techniques, and the use of synthetic captions generated with vision-language models such as CogVLM \cite{esser2024scalingrectifiedflowtransformers}. While the source data may remain similar in origin—likely large-scale web-scraped corpora—the enhanced preprocessing pipeline alters how visual concepts are represented and associated in the model’s latent space.

These differences in training pipelines can affect how each model encodes relationships between facial characteristics, perceived attractiveness, race, gender, and other high-level attributes. 
Our analysis examines potential divergences between the outputs of SD 2.1 and SD 3.5 Medium in light of these underlying differences, with particular attention to their implications for demographic representation and aesthetic normativity. While our quantitative analyses (Sections \ref{sec:rq1.1}–\ref{sec:rq1.2}) measure these differences through embedding space calculations, our qualitative visual assessment (Section \ref{sec:qualitative_analysis}) complements them by examining observable aesthetic patterns in the generated imagery.
\section{Gender Classification Performance as a Function of Different Attributes}
\label{appendix:classification}
\begin{figure}
    \centering
    \includegraphics[width=\linewidth]{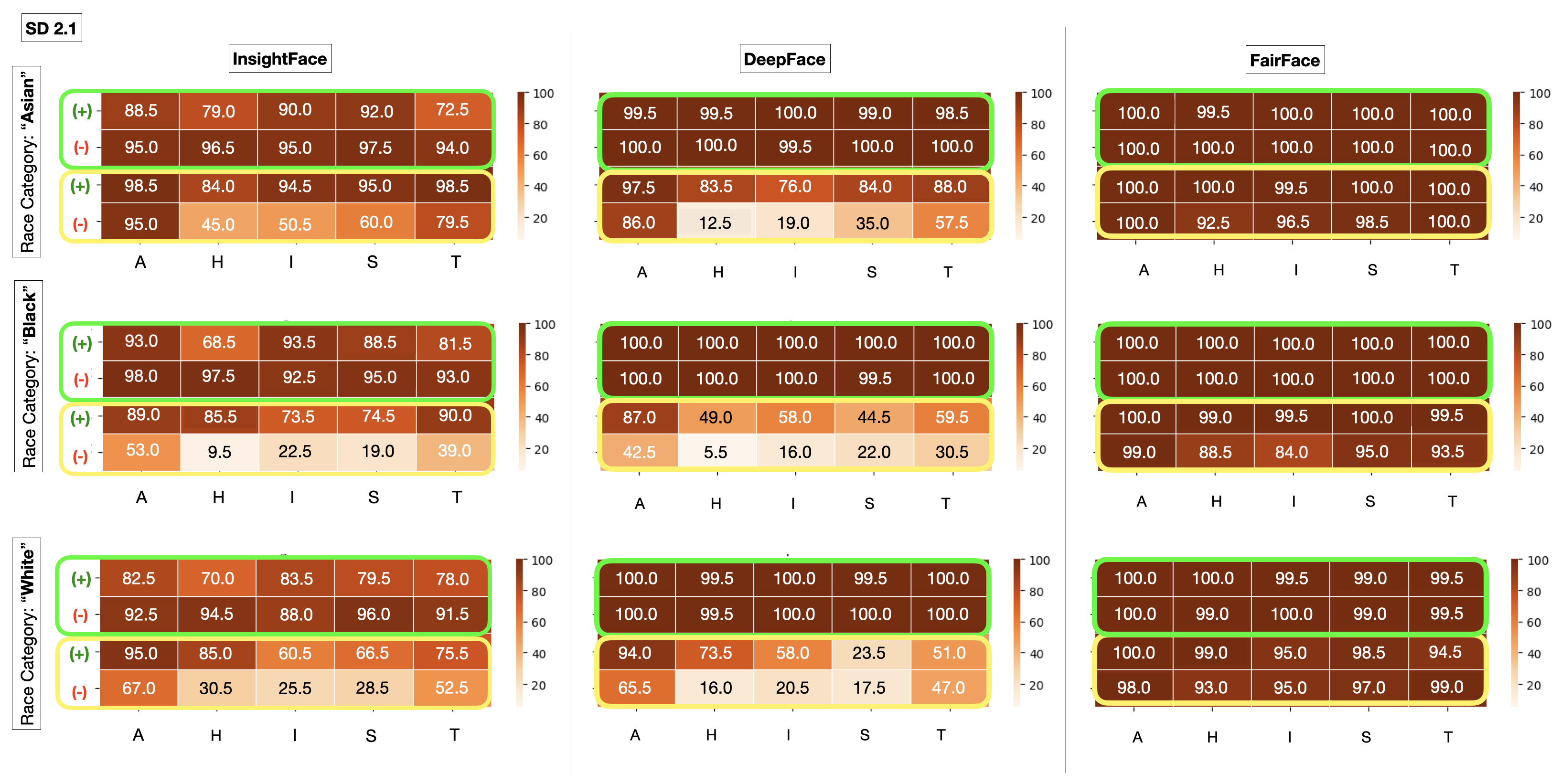}
    \caption{Heatmaps of gender classification accuracy (gender x race) (\textbf{SD 2.1}) for InsightFace, DeepFace, and FairFace. A = Attractiveness, H = Happiness, I = Intelligence, S = Sociability, T = Trustworthiness. Women = Yellow \colorsquare{c_female}, Men = Green \colorsquare{c_male}.
    \underline{Neutral face accuracies} (\%) \textbf{InsightFace:}  (Asian W)  90.0, (Black W) 39.5, (White W) 57.0 - (Asian M) 86.0, (Black M) 90.0, (White M) 89.5; 
    \textbf{DeepFace:} (Asian W)  61.0, (Black W) 32.5, (White W) 56.0 - (Asian M) 100, (Black M) 99.5, (White M) 100; 
    \textbf{FairFace:} (Asian W)  100, (Black W) 95.5, (White W) 95.0 - (Asian M) 100, (Black M) 100, (White M) 100.
    }

    \label{fig:sd21_gender_complete}
\end{figure}

\begin{figure}
    \centering
    \includegraphics[width=\linewidth]{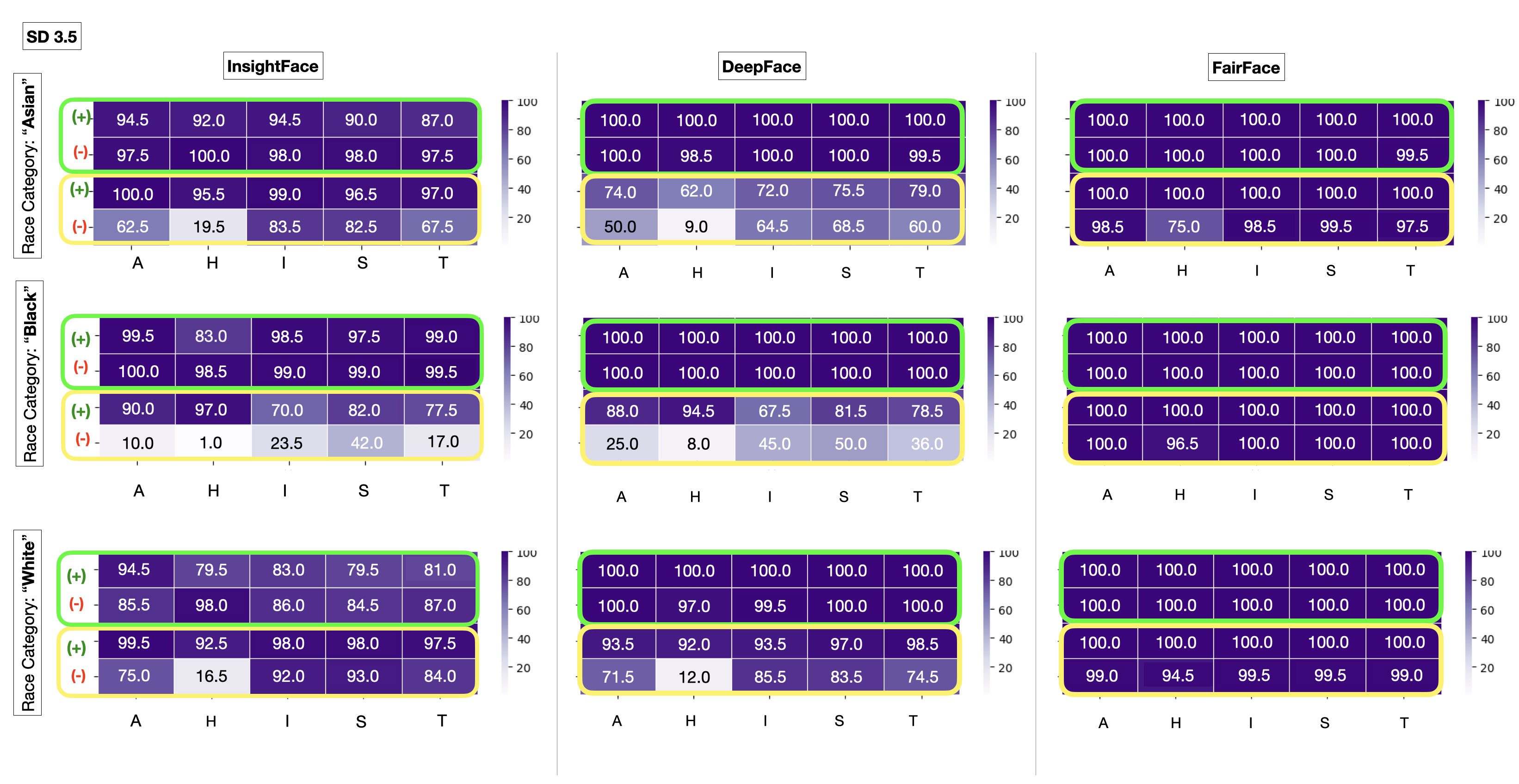}
    \caption{Heatmaps of gender classification accuracy (gender x race) (SD 3.5) for InsightFace, DeepFace, and FairFace. A = Attractiveness, H = Happiness, I = Intelligence, S = Sociability, T = Trustworthiness. Women = Yellow \colorsquare{c_female}, Men = Green \colorsquare{c_male}.
    \underline{Neutral face accuracies} (\%) \textbf{InsightFace:} (Asian W)  98.0, (Black W) 58.5, (White W) 92.5 - (Asian M) 89.5, (Black M) 97.0, (White M) 83.0; 
    \textbf{DeepFace: }(Asian W)   77.5, (Black w) 58.5, (White W) 93.5 - (Asian M) 100, (Black M) 100, (White M) 100; 
    \textbf{FairFace:} (Asian W)  100, (Black W) 99.5, (White W) 100 - (Asian M) 100, (Black M) 100, (White M) 100.
    }

    \label{fig:sd35_gender_complete}
\end{figure}

To complement the main results presented in Section~\ref{sec:classification}, this appendix reports the disaggregated gender classification performance across racial categories for both SD 2.1  (Figure \ref{fig:sd21_gender_complete}) and SD 3.5 (Figure \ref{fig:sd35_gender_complete}) generated images. While the main text aggregates performance by gender for clarity and space constraints, the disaggregated analysis reveals important patterns related to intersectional disparities.

As shown in the following tables and figures, the classification performance for female faces varies more substantially across racial groups than for male faces, which remain consistently well-classified. This confirms that algorithmic bias in gender classification is not only gendered but also racialized, affecting specific subgroups more severely.

\section{Happy-Beauty Conflation}
\begin{figure}[h]
    \centering
    \includegraphics[width=0.65\linewidth]{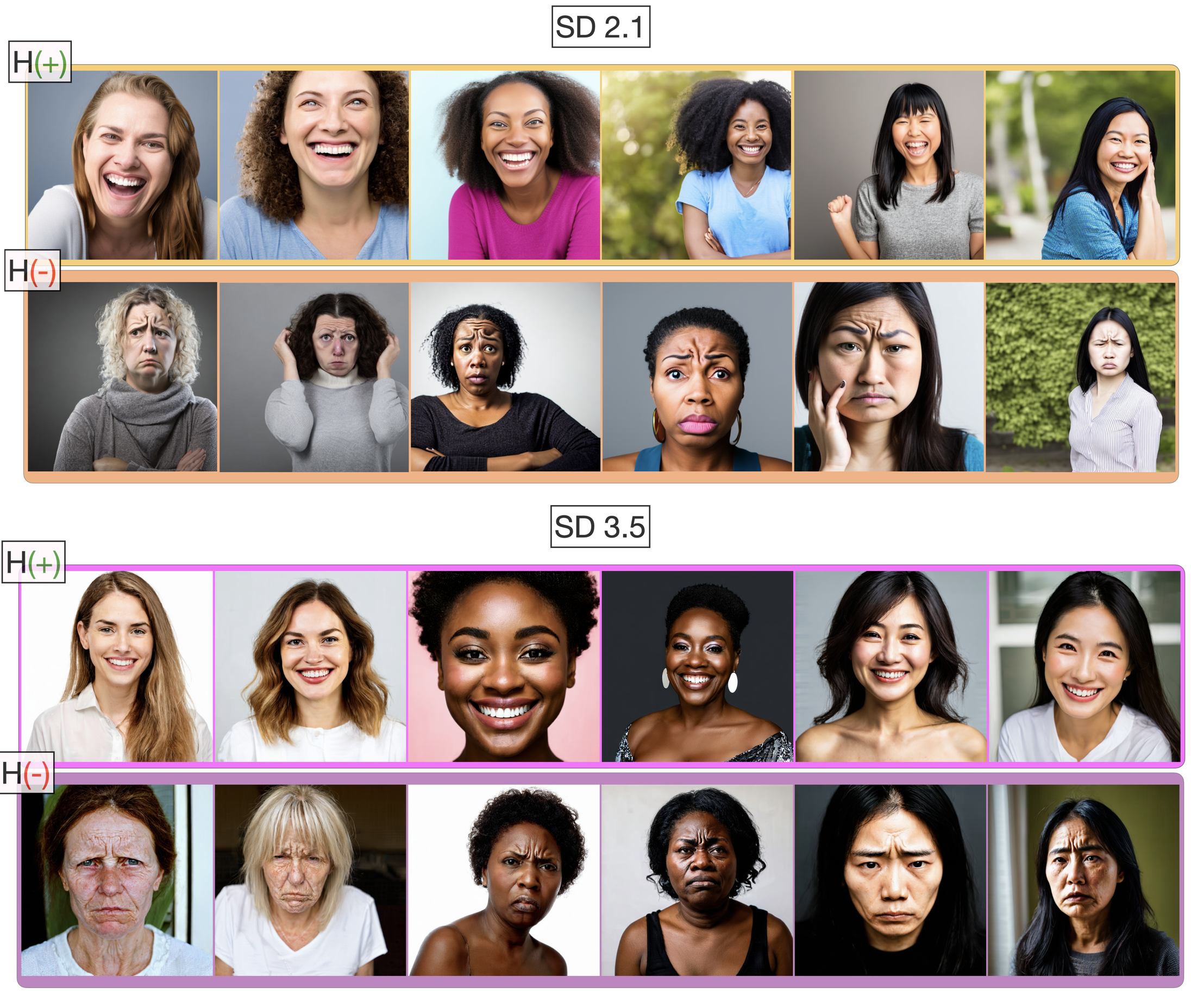}
    \caption{Examples of generated images for "Happy" and "Unhappy" women for SD 2.1 (orange) and SD 3.5 (purple). }

    \label{fig:tobehappywoman}
\end{figure}
\begin{figure}[h]
    \centering
    \includegraphics[width=0.65\linewidth]{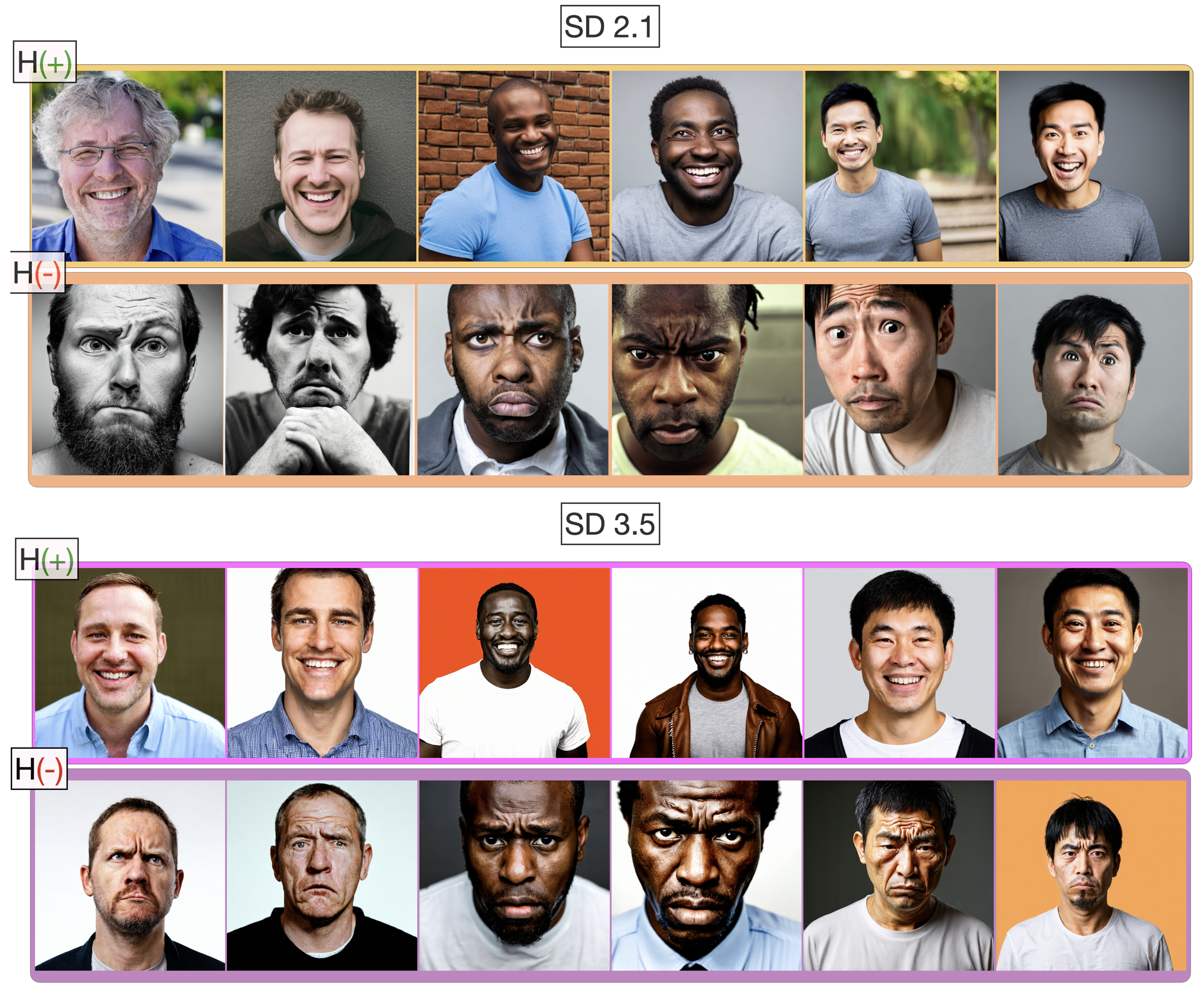}
    \caption{Examples of generated images for "Happy" and "Unhappy" men for SD 2.1 (orange) and SD 3.5 (purple). }

    \label{fig:tobehappyman}
\end{figure}
\section{Gendered Exposure): NSFW Content Analysis}
\label{app:nsfw_analysis}
\begin{figure}[h]
    \centering
    \includegraphics[width=0.9\linewidth]{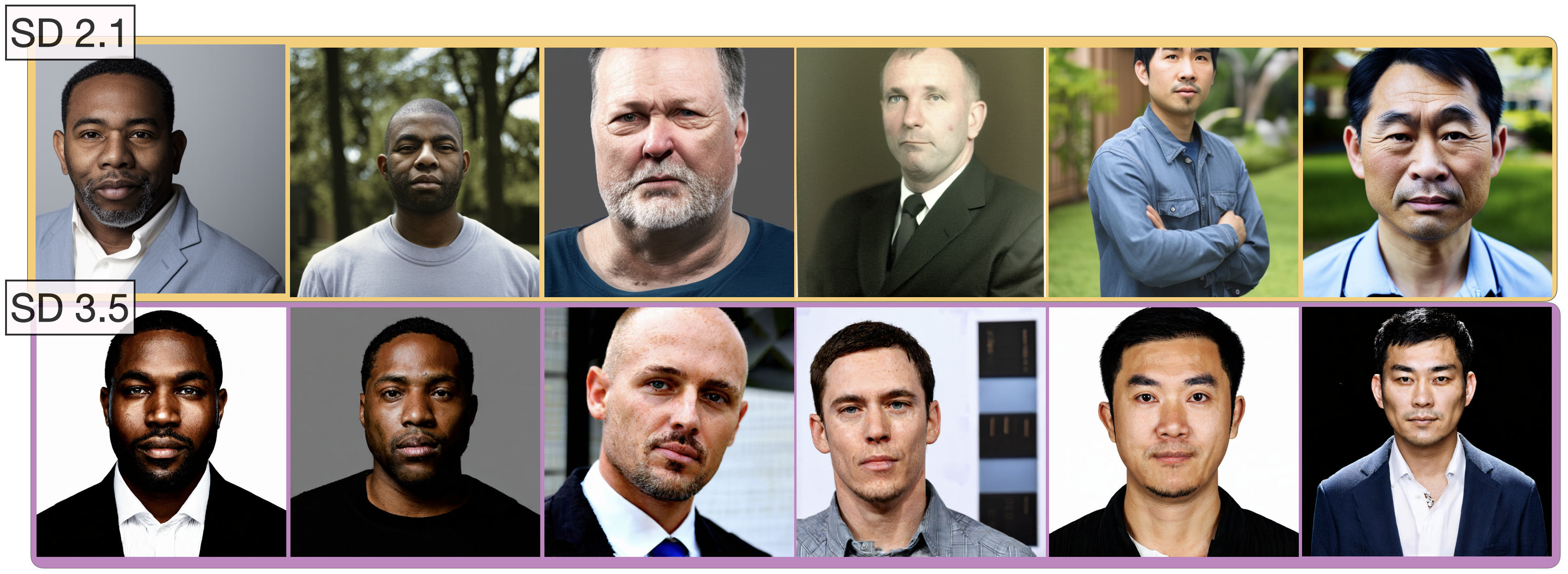}
    \caption{Examples of generated images for "Neutral" men for SD 2.1 (orange) and SD 3.5 (purple). }

    \label{fig:men_Exp}
\end{figure}

To quantitatively assess patterns of sexualized representation observed in our visual analysis—particularly the recurrent exposure of skin in generated images of women—we conducted a supplementary analysis using NSFW (Not Safe For Work) content detection. This analysis was prompted by a consistent pattern across both SD 2.1 and SD 3.5: female-presenting figures were more frequently depicted with visible necklines, shoulders, and cleavage, irrespective of the specified attribute (see Section~\ref{sec:qualitative_analysis} and Figure~\ref{fig:tobehappy}.(a)). 
\begin{figure}[h]
    \centering
    \includegraphics[width=0.9\linewidth]{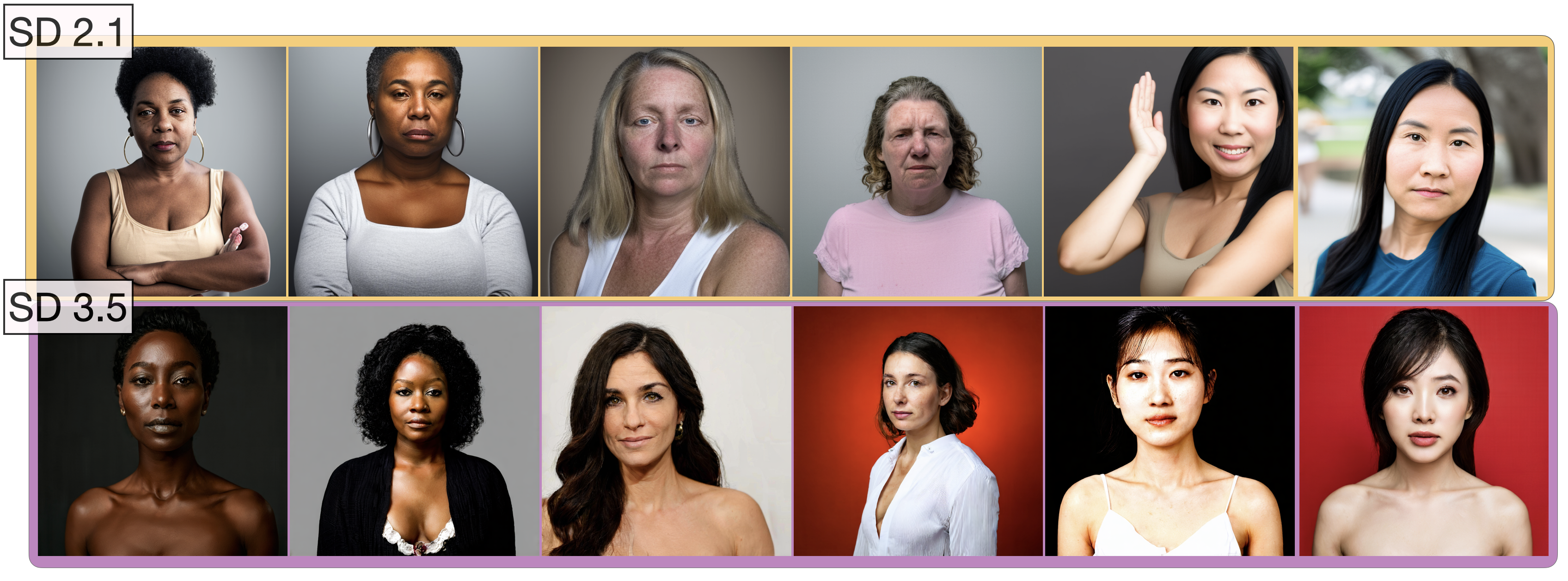}
    \caption{Examples of generated images for "Neutral" women for SD 2.1 (orange) and SD 3.5 (purple).}

    \label{fig:women_Exp}
\end{figure}
Building on prior work demonstrating the sexualization of women in AI-generated imagery \cite{wolfe2023contrastive, washingtonpost2023ai, guilbeault2024online, lan2025imaginingfareastexploring}, and evidence of pornographic bias in training data—such as the fact that 20\% of captions for “Latina” in Stable Diffusion’s training corpus included explicit terms \cite{bloomberg2022ai}—our aim was to determine whether these qualitative observations were reflected in NSFW detection rates.

We applied a standard NSFW detection algorithm \cite{laborde2018nsfw} to uncropped versions of all generated images, comparing detection rates across attributes, gender, and race. While we acknowledge that NSFW detectors themselves may be biased—tending to flag more images of women than men as sexually suggestive \cite{mauro2023biased}—the analysis nonetheless offers insight into how generative models encode gendered exposure as a latent visual feature.

\begin{figure}
     \centering
     \includegraphics[width=0.8\linewidth]{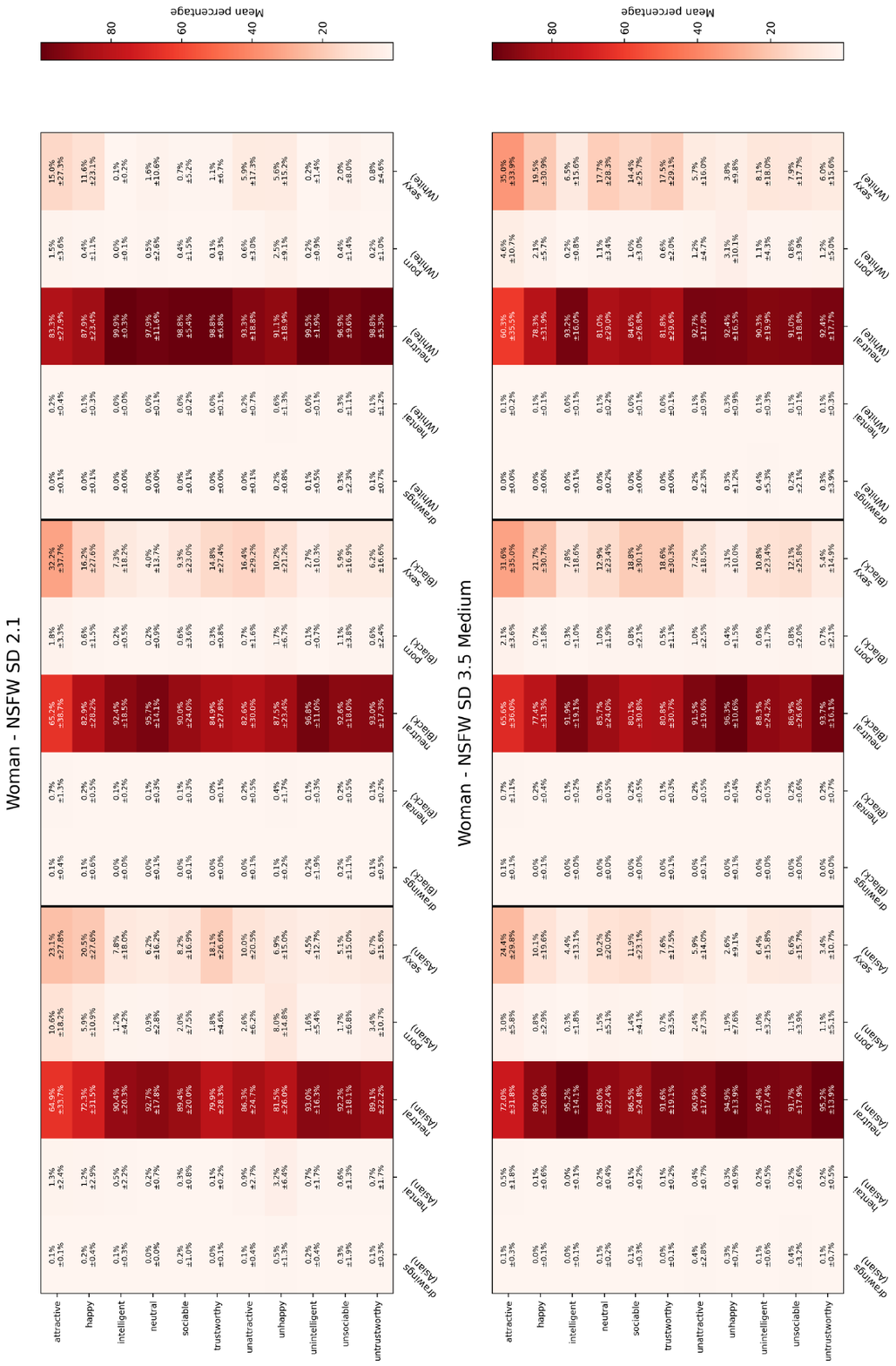}
     \caption{Heatmaps showing NSFW detection rates (\%) for generated images of women across different attributes and demographic groups. The top panel  displays results for SD 2.1, while the bottom panel displays results for SD 3.5 Medium.}

     \label{fig:complete}
 \end{figure}

\begin{figure}
     \centering
     \includegraphics[width=0.8\linewidth]{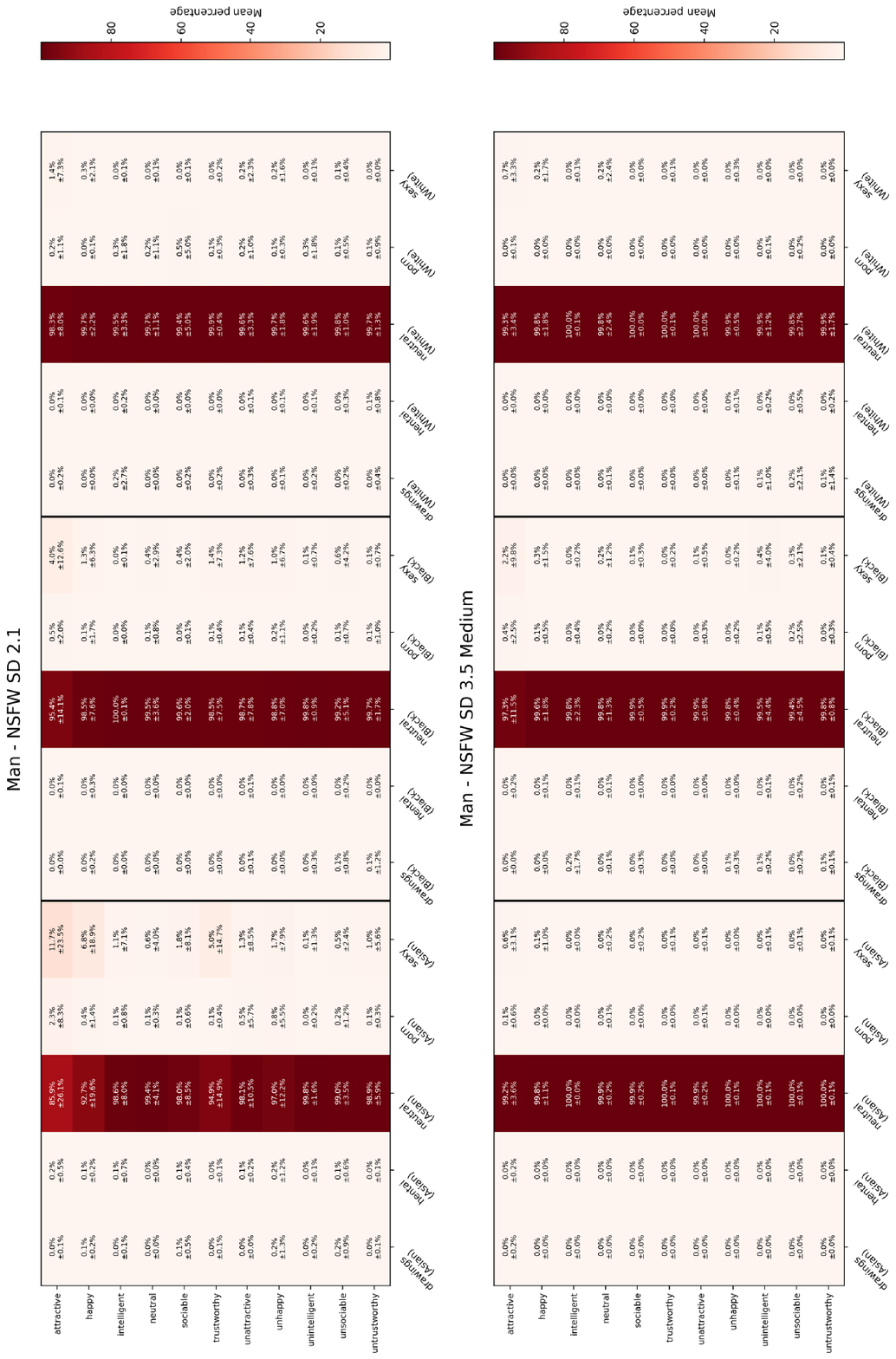}
     \caption{Heatmaps showing NSFW detection rates (\%) for generated images of men across different attributes and demographic groups. The top panel  displays results for SD 2.1, while the bottom panel displays results for SD 3.5 Medium.}

     \label{fig:complete_men}
 \end{figure}

\end{document}